\documentclass[runningheads]{llncs}
\usepackage{graphicx}
\usepackage{subfiles}
\usepackage{amssymb,mathabx,wasysym}

\usepackage{soul}

\usepackage[normalem]{ulem}

\usepackage{array,bold-extra,tikz}
\usetikzlibrary{automata,arrows,shapes,snakes,topaths,trees,backgrounds,shadows,matrix,positioning,calc}
\usepackage{rotating}
\usetikzlibrary{arrows.meta}
\usepackage{xcolor}
\usepackage[colorinlistoftodos]{todonotes}
\usepackage{latexsym,stmaryrd} 

\tikzset{
	mystyle/.style={
		rectangle,
		inner sep=0pt,
		text width=3.6cm,
		minimum height=1.6cm,
		align=center,
		draw=black,
		fill=white
	}
}

\tikzset{
	mystyle1/.style={
		rectangle,
		inner sep=0pt,
		text width=4cm,
		minimum height=1.8cm,
		align=center,
		draw=black,
		fill=white
	}
}

\tikzset{
	mystyle2/.style={
		rectangle,
		inner sep=0pt,
		text width=4.2cm,
		minimum height=2.4cm,
		align=center,
		draw=black,
		fill=white
	}
}

\tikzset{
	mystyle3/.style={
		rectangle,
		inner sep=0pt,
		text width=5.4cm,
		minimum height=1.2cm,
		align=center,
		draw=black,
		fill=white
	}
}

\tikzset{
	mystyle4/.style={
		rectangle,
		inner sep=0pt,
		text width=3cm,
		minimum height=1.2cm,
		align=center,
		draw=black,
		fill=white
	}
}

\begin{document}
\title{Towards AI Logic for Social Reasoning}

%
\author{Huimin Dong\inst{1} 
\and
R\'{e}ka Markovich\inst{2}
\and
Leendert van der Torre\inst{1,2}
}
%
%
\institute{Department of Philosophy, Zhejiang University, China \and
Department of Computer Science, University of Luxembourg, Luxembourg
\email{
huimin.dong@xixilogic.org,
reka.markovich@uni.lu, 
leon.vandertorre@uni.lu,
}}
\maketitle              
\begin{abstract}
Artificial Intelligence (AI) logic formalizes the reasoning of intelligent agents. In this paper, we discuss how an {\em argumentation-based} AI logic 
could be used also to formalize important aspects of {\em social} reasoning. Besides reasoning about the knowledge and actions of individual agents,  social AI logic can reason also about social dependencies among agents using the rights, obligations and permissions of the agents. We discuss four aspects of social AI logic.
First, we discuss how {\em rights} represent relations between the obligations and permissions of intelligent agents. Second, we discuss how to argue about the {\em right-to-know}, a central issue in the recent discussion of privacy and ethics. Third, we discuss how a wide variety of {\em conflicts among intelligent agents} can be identified and (sometimes) resolved by comparing formal arguments.  Importantly, to cover a wide range of arguments occurring in daily life, also fallacious arguments can be represented and reasoned about. Fourth, we discuss how to argue about the {\em freedom to act} for intelligent agents. Examples from social, legal and ethical reasoning highlight the challenges in developing social AI logic. 
The discussion of the four challenges leads to a research program for  {\em argumentation-based social AI logic}, contributing  towards the future development of {\em AI logic}.
\end{abstract}
%
%
%


\section{Introduction}



Artificial Intelligence (AI) is not only a sub-discipline of computer science, but it also overlaps with other disciplines such as philosophy, psychology, sociology, law, ethics, and economics. AI researchers build programs mimicking human intelligence in order to enable computers displaying so-called intelligent behaviour~\cite{russell2016artificial}, and they develop various kinds of techniques for perception, representation and reasoning, learning, natural interaction, and societal impact of AI.
Logic can be used in various of these AI fields~\cite{sep-logic-ai,gabbay1994handbook}, in particular in the context of explainable AI to complement black box technologies~\cite{hollander2011current}.

With the rise of artificial intelligence and machine learning, questions are raised about the reasoning of autonomous robots and other intelligent systems about their own actions, but also about the balance their own desires and goals with the social, legal, and ethical norms of society~\cite{normas}. Moreover, questions are raised how intelligent agents take the actions of other agents into account~\cite{dastani2005decide,thomason2014formalization,mccarthy1980circumscription}. For example, an autonomous car must not only be able to drive and find its way, but it also needs to take into account traffic law, and it needs to reason about the behavior of other cars, either autonomous or driven by humans.
In this paper, we therefore address the following challenge.

\begin{quote}
Research question: how can AI logic  be extended to social reasoning?
\end{quote}

Though some steps towards the development of social AI logic have already been made, for example in deontic logic, we would like to emphasize that the development of social AI logic is an ambitious research program. At an abstract level, the main difference between social and practical reasoning with classical reasoning, is that agents may disagree about the right action to take, or even about the state of the world~\cite{caldas1999origin,flentge2001modelling,hales2002group}. Formal approaches to practical reasoning aim to capture the many ways in which individual and collective concepts interact, such as knowledge-based obligation~\cite{pacuit2006logic}. In a sense, social collective behavior is the sum of the individual behaviors of the agents~\cite{horty2001agency}, but it is still challenging to relate social reasoning to the practical reasoning of the agents in formal theories. Individual agent reasoning is built on ability and knowledge, where the abilities of an agent are reflected by the actions it can execute, and its knowledge contains also the uncertainty, under which it makes decisions~\cite{horty2019epistemic,boella2007game}.
In particular, we address the following questions concerning the development of social AI logic:

\begin{enumerate}
\item
How do {\em rights} represent relations between the obligations and permissions of intelligent agents?
\item
How to argue about claims combining both AI and social concepts like the {\em right to know}?
\item
Which kind of {\em conflicts among intelligent agents} can be identified, and when and how can they be resolved?
\item
How to argue about the {\em freedom to act} for intelligent agents?
\end{enumerate}

To answer the first question, we introduce rights and permissions in AI logic as our central concepts for social reasoning. We explain how they interact with prohibitions and obligations in different ways, 
and how they relate to other social concepts like freedom to act~\cite{hansson13:_variet_permis,dong2015three,dong2019aspic}.
Rights and permissions emphasize how social norms empower agents to achieve things they could not achieve by themselves~\cite{sergot13_normative,deonhandbook,normas}. From the perspective of intelligent agents, prohibition and obligation are more negative in the sense that they emphasize how social norms constrain our individual planning~\cite{deonhandbook}. 

To answer the second question, we focus on the right-to-know, because it is understudied, despite its practical implications for important issues such as privacy \cite{privacy}, but also because the right to know can concern several different things and, therefore, many different social relations: from patients' right to know about their own medical results and condition, the soldiers' right to know the cause they are fighting for \cite{rtk}, the citizens' right to know about their governments' doings, until the US environmental law declaring the people right to know what chemicals they are exposed to.

In the third question, the omnipresence of conflict and uncertainty in social reasoning leads to the use of formal argumentation~\cite{dong2019aspic}. Conflict is not ignored but accepted, and the aim of reasoning is often not only to resolve such conflicts, but also to agree where the agents disagree.

In the discussion of the fourth and final question, we consider freedom in the context of rights, but also in the context of permission. We consider challenges concerning free-choice permission discussed in the literature, and show how there challenges can be addressed using formal argumentation.

From our  answers to the four research questions, we extract a research program for the future development of {\em argumentation-based social AI logic}, or more generally, social AI logic. The challenges in developing social AI logic in this paper are illustrated by 
examples from social, legal and ethical reasoning. Here we like to make the following well-known observations about modelling real life scenarios, either in logic or in some other knowledge representation language.
First, the problem of formalization is hard, it is a skill, and some people call it an art. 
Second, often there is no ``best'' formalization, and typically there are several formalizations of a claim, each with their advantages and disadvantages.
Third, the key challenge of modelling examples concerns the development of a modelling methodology. It is like in decision theory: the first guess of utility function and probability distribution is always wrong. People need to play with it (in the case of decision theory, with lotteries) to better learn themselves. It leads to the field of preference elicitation in decision analysis. 
However, a further discussion of the modelling methodology is outside the scope of this article.

The layout of this paper is as follows.
In Section 2 we introduce the language of AI logic we use in this paper, and in particular we explain how rights represent dependencies between agents by relating the obligations and permissions of agents to the obligations and permissions of other agents. In Section 3 we show how formal argumentation can be used to represent and resolve (some) conflicts among agents. In Section 4, we show how to express the freedom of agents to choose, and how this is reflected in the choice between so-called strong and weak permission. 
In Section 5 we introduce our research program for argumentation-based social AI logic.

\section{Rights and Permissions in AI logic}






AI logic formalizes the practical reasoning of intelligent autonomous agents. In this paper, we use an AI logic that has two distinctive features with respect to classical logic. In this section, we explain how  AI logic extends classical logic with modal operators for actions, beliefs, knowledge, right, obligations and permissions of intelligent autonomous agents. 
In the following section, we explain how AI logic uses techniques from formal argumentation to represent common-sense reasoning, and deals with uncertainty, conflicts and exceptions. 
We use the following running example in this section.

\begin{example}[Sensitive data scenario] \label{exam:data}
	Someone's health data counts as sensitive data and as such is subject to strong protection principles in most of the countries (in the European Union to so-called GDPR, Regulation (EU) 2016/679 of the European Parliament and of the Council of 27 April 2016 on the protection of natural persons with regard to the processing of personal data and on the free movement of such data, and repealing Directive 95/46/EC). This means that others are not allowed to know these data. However, if someone is ill and in need of a treatment, we would all agree that the doctors have to provide this treatment. But fulfilling this obligation of them requires that they know the data regarding the patient's health. 
\end{example}

Various claims from the example scenario above are visualized in Figure 1 below, together with a formalization in AI logic. Moreover, the claims are grouped in two arguments visualized using vertical arrows. The four claims on the left constitute the argument with conclusion that the doctor is not permitted to know the sensitive information, and the four claims to the right constitute the argument that the doctor is permitted to know that. The top claims are the conclusions of the argument, and the other claims are supporting those conclusions. Moreover, the fact that the doctor's permission to know prevails is modelled by the arrow from the top right box to the left top box. In this section we consider only the natural language statements and their formalization in AI logic. The argumentation aspects are discussed in more detail in Section 3.


\begin{figure}
	\noindent
	\makebox[\textwidth]{
	\begin{tikzpicture}[scale = .8]

	\node[mystyle1] (A) at (-5,-1.5) {\scriptsize * The patient has a right to privacy\\ $A_1 = R_p(privacy)$};

	\node[mystyle1] (B) at (-5,2) {\scriptsize Others are not allowed to know about his sensitive data. \\ $A_2 = A_1 \mapsto \bigwedge_{a \in Agt/\{p\}} \neg PK_a(sensitive)$
	};

	\node[mystyle1] (C) at (1,2) {\scriptsize * Data regarding someone's health is sensitive data. \\ $A_3 = \Box(illness \rightarrow sensitive)$};

	\node[mystyle1] (D) at (-2,6) {\scriptsize The doctor is not permitted to know about the patient's illness. \\ $A_4 = A_2, A_3 \Mapsto \neg PK_d(illness)$ 
	};


	\node[mystyle1] (A') at (4,-1.5) {\scriptsize * The doctor has an obligation to treat an ill person. \\ $B_1 = O_d \langle d\rangle(treat)$
	};
	
	\node[mystyle3] (B') at (8,2) {\scriptsize * The doctor cannot treat a person if she doesn't know about the illness. \\ $B_2 = \Diamond\langle d \rangle(treat) \rightarrow K_d(illness)$
	 };
	
	\node[mystyle1] (C') at (6,6) {\scriptsize The doctor is permitted to know about the patient's illness. \\ $B_3 = B_1, B_2 \mapsto PK_d(illness)$};


    \draw[<-] (B) --  (A) node[draw=none,fill=none,font=\scriptsize,midway,right] {};
	
    \draw[dashed] (B) --  (-5, 4) -- (1, 4) -- (C);

    \draw[dashed, ->] (-2, 4) -- (D) ; 


    \draw (A') --  (4, 4) -- (8, 4) --  (B') node[draw=none,fill=none,font=\scriptsize,midway,right] { };

     \draw[ ->] (6, 4) -- (C') ;

     \draw[postaction={draw,line width=1.5pt,solid}, <-] (D) -- (C');
    		
	\end{tikzpicture}
}
	\caption{Two arguments in the sensitive data scenario, each containing four claims. } \label{fig:doctor}

\end{figure}
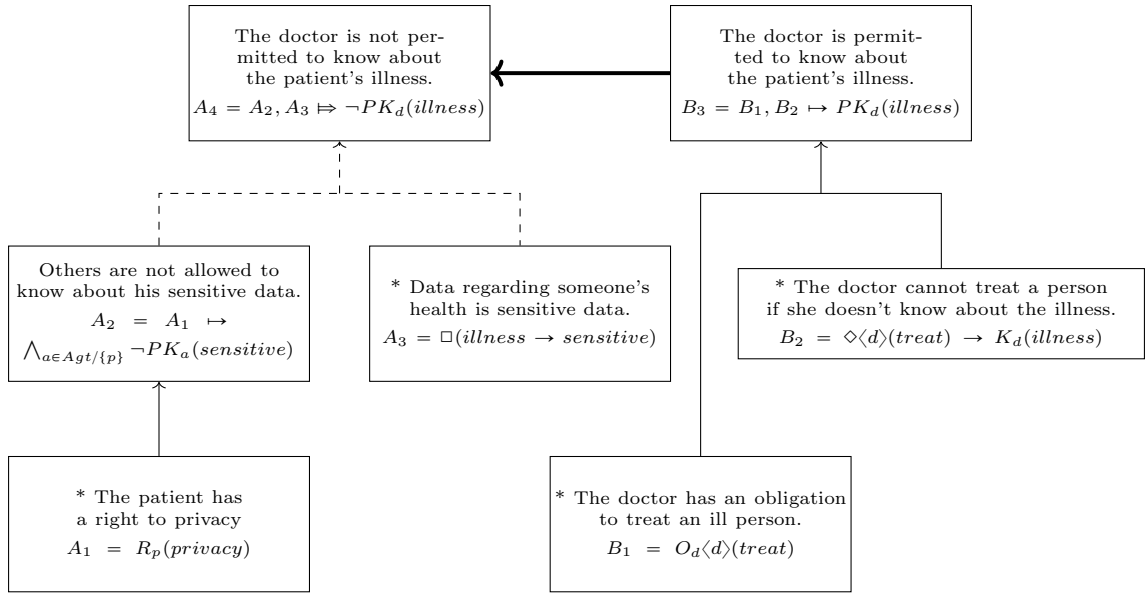

\subsection{Language of modal AI logic}

Consider a classical logic language with formulas like {\it illness} and {\it treat}, to represent that the patient is ill and the patient has been treated. We can also add variables, such that {\it illness(patient)} represents that the patient is ill. Moreover, considering agents like {\it doctor, patient} and {\it Suzy}, in modal logic we can have formulas like $K_{\mbox{\it doctor}}$({\it illness}) to express that the doctor knows that the patient is ill, and $\neg K_{\mbox{\it patient}}$({\it illness}) to express that the patient does not know that he or she is ill. Here we may adopt the usual assumption in epistemic logic that beliefs may be mistaken but knowledge must be correct. In other words, when a doctor knows that the patient is ill, then this implies that the patient is ill. For example, knowledge may be modelled as justified, true belief.

We consider social aspects like rights, obligations and permissions. For example, $O_{\mbox{\it doctor}}$({\it treat}) may express that for the doctor it is obliged that the patient is treated, and $P_{\mbox{\it patient}}$({\it illness}) may express that the patient is permitted to be ill. In some cases we can also index the modal operators with two agents in order to express that the obligation is directed (see Section 2.2). For example $O_{\mbox{\it doctor,patient}}$({\it treat}) may express that for the doctor it is obligatory with respect to the patient that the patient is treated. That is, this is a duty of the doctor toward the patient. In general, there are several ways in which an example can be formalized, and this is left to the person modelling the example.

Practical reasoning is about what agents do, so also actions must be represented in the language~\cite{horty2001agency}. For example $[{\it doctor}] ({\it treat})$ stands for the action that the doctor sees to it that the patient is treated. Finally, we use a universal modality $\Box$ to represent necessities, for example $\Box(illness\rightarrow sensitive)$ expresses that data regarding someone's health is sensitive data.

The advantage of modal logic is that all the expressions can be mixed in various ways, and this will be used extensively in this paper. For example $P_{\mbox{\it doctor}}[doctor]K_{\mbox{\it patient}}$({\it illness}) represents that for the doctor it is permitted that the he see to it that the patient knows about her illness. We can also represent universal permissions and obligations by leaving out the agents. An example is $P K_{\mbox{\it doctor}}$({\it illness}).

\subsection{Rights representing social dependencies} \label{sec:right}

\begin{figure}
	\noindent
	\makebox[\textwidth]{
\begin{tikzpicture}[scale=3, >=stealth, line width=.5mm]
\begin{scope}[]
\node (v1) at (0,0) {duty};
\node (v2) at (0,1) {claim-right};
\node (v3) at (1,1) {freedom};
\node (v4) at (1,0) {no-claim};
\draw[<->]  (v1) -- (v2) node[midway, above, sloped]{\footnotesize correlatives};
\draw[<->]  (v3) -- (v4) node[midway, below, sloped, rotate=180]{\footnotesize correlatives};
\draw[<->, blue]  (v2) -- (v4);
\draw[<->, blue]  (v1) -- (v3);
\node at (0.5,0.75) {\footnotesize opposites};
\end{scope}
\begin{scope}[xshift=2cm]
\node (v1) at (0,0) {liability};
\node (v2) at (0,1) {power};
\node (v3) at (1,1) {immunity};
\node (v4) at (1,0) {disability};
\draw[<->]  (v1) -- (v2) node[midway, above, sloped]{\footnotesize correlatives};
\draw[<->]  (v3) -- (v4) node[midway, below, sloped, rotate=180]{\footnotesize correlatives};
\draw[<->, blue]  (v2) -- (v4);
\draw[<->, blue]  (v1) -- (v3);
\node at (0.5,0.75) {\footnotesize opposites};
\end{scope}
\end{tikzpicture}
}
\caption{Representation of Hohfeldian Rights} \label{fig:hohfeld}
\end{figure}
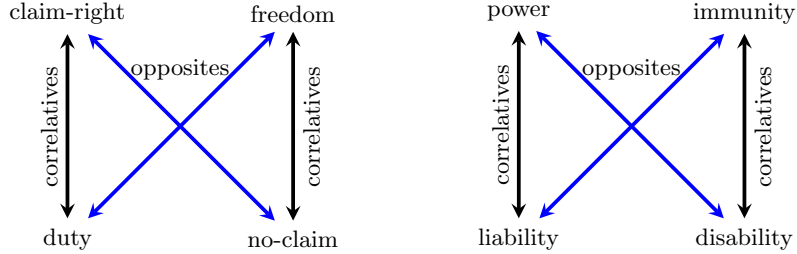

Rights play a central role in this paper, since they point out a crucial social phenomenon: how the agents' social (in the specific context: normative) positions depend on each-other and each-other's positions. In legal practice it is well known that talking about ``rights'' in itself is ambiguous. This ambiguity easily leads to a conceptual obscurity, thus, a hundred year ago, an American legal theorist differentiated between the possible meanings \cite{Hohfeld,MarkovichSL}. In the system of Hohfeld, which consists of four different right-relations, there are four different rights, each comes with a given type of duty on the other side: someone's right always means \textit{someone else's} duty, he calls these pairs, that is, \textit{normative positions}~\cite{sergot13_normative} at the ends of the four possible relations, correlatives, as they always come together (which, from a logical point of view, will be equivalence). The system Hohfeld drew can be graphically represented in Figure~\ref{fig:hohfeld} (rights in the upper row, duties in the lower one).

These positions emerge most visibly in case of a contract: the seller's (claim-)right to the purchase-price obviously means the buyer's duty to give it to her. But this phenomenon is way more general. Regarding the epistemic positions, if someone has a (claim-)right to know something, it means that the other agent has a duty directed toward the previous one to tell him: $R_a[b]K_a\varphi \Leftrightarrow O_{b,a}[b]K_a\varphi$. If someone has a freedom to know something, though, that only means that the other agent has no (claim-)right toward him that he doesn't (get to) know. We usually consider this normative position as 
\textit{permission}. This will be analyzed in the next subsection.
 
The square on the right-hand side exhibits a very similar structure, those positions, though, are dynamic~\cite{kanger1966rights,lindahl1977position,dong2017dynamic}, they are about the \textit{potential} to change other's rights and duties~\cite{MarkovichSL}. For example, if we mean the right to know something as a power, that means that the agent having this power can bring about a duty directed on the other agent---whose position is called liability in this system---to let her know: $Power_{a,b}[a]O_{b,a}[b]K_a\varphi$. Meanwhile, if someone has an immunity regarding her knowledge, we would mean that the other agent has no power to generate a duty on her to tell him. 
Applying and relying on Hohfeld's system and, therefore, on agent-indexed normative positions doesn't exclude talking about general (from the legal point of view, so-called absolute) rights and duties. For instance the often discussed civil liberties, as the right to free speech means that the agent has a claim-right towards everyone else to let her express whatever she wants \cite{MarkovichSL}. Just like we have an obligation not to kill, which, in general, doesn't need to indicate toward whom this duty exist because it exists toward everyone else: $O_a\neg[a](kill)$ (and this covers the conjunction of each rights-relation between us and the other members of the society \cite{dong2020multi,MarkovichIFCoLogRP}). What is more, if we consider this obligation as being general in the sense of being valid for everyone, focusing only on the result that no one should be killed, we might say: $O\neg(kill)$.

\subsection{Free-choice permission} \label{sec:permission}




Handling the notion of permission in logic on a systematic level started at the same time as handling obligation: with creating the so-called deontic logic \cite{deonhandbook}, which we usually ascribe to Georg von Wright \cite{vonWright1951}.
Von Wright's sense of weak permission, usually, is understood as ``the lack of prohibition''. So the expression ``it is permitted'' is supposed to mean``it is not forbidden''. This reductive account simplifies weak permission as the negation of forbiddance, or, in other words, the dual of obligation. This interdefinability is one of two basic axioms in standard deontic logic~\cite{mcnamara2006deontic}. Therefore, weak permission adopts the \emph{upward} principle as the possibility modality does in normal modal logic: ``If it is  permitted to check the patient's medical record and treat him, then it is permitted to treat the patient'' $P_{doctor}(record \wedge treat) \rightarrow P_{doctor}(treat)$.

Yet it is inevitable that the feature of closed \emph{upward} let weak permission have too strong derivation power than we need in social reasoning. Weak permission brings us back to a variant of Ross' paradox~ \cite{mcnamara2006deontic}. Given that it is permitted to check the patient's medication record, it then infers that it is permitted to check the patient's medication record or sell it to the insurance company $P_{doctor}(record) \rightarrow P_{doctor}(record \vee sell)$~\footnote{Since, as we mentioned above, a permission always concerns the agent's action whose permission we talk about $P_{doctor}(record)$ or $P[d](record)$ are shorter versions of writing that $P_{doctor}[doctor](record)$.}. But the latter is not ethically or legally allowed to do.


Von Wright later said, ``(Strong) [p]ermission (...) means freedom to choose between all the alternatives, if any, covered by the permitted thing''~\cite[p.32]{von1968essay}. This ``open'' flavor of ``freedome to chose'' is well known in philosophy as permission \emph{at liberty}~\cite{raz1975permissions}, and also commonly used in ordinary language and legal context: 
	``Exactly how much to tip a server is at the discretion of the customer'',  
	``Bail is granted at the discretion of the court''.
In contrast to weak permission, the open sense is \emph{downward} closed but restricted (FCP): ``If it is permitted to know the patient's medical record, then it is permitted to know the patient's illness'' $PK_{doctor}(record) \rightarrow PK_{doctor}(illness)$, as that the situation of the illness is a type of medical record $\Box(illness \rightarrow record)$. This kind of permission has been developed by van Benthem in modal logic~\cite{van1979minimal}, in which strong permission not only has to follow the axiom in this open sense, but also have to obey the axiom called ``obligation as the weakest permission'' (OWP)~\cite{ano2015symbolic}. The latter gives the ``non-violation of norm'' restriction for strong permission: All strong permissions have to stay inside the zone set up by obligations. While the doctor may know and use the patinet's e-mail address to communicate with him, it is forbidden to misuse the patient's data, for example, she cannot send spam to him $O_{doctor}\neg(spam)$. Though this permission is for the doctor to know and use the patient's e-mail address $PK_{doctor}(email)$, for the doctor (normally) what is permitted is restricted by the prohibition $O\neg_{doctor}(email \rightarrow \neg (spam))$. This is similar to von Wright's viewpoint shown beforehead~\cite{vonWright1963}.  

Some may argue that the derivation power of strong permission is still too strong and thus faces some difficulty in social reasoning, given the free choice permission paradox~\cite{hansson13:_variet_permis}. The worry concerns that \emph{downward} closure can be led given some non-normal case: Permitting the doctor to check the patient medical record, then the doctor is permitted to know the record and destroy it $PK_{doctor}record \rightarrow P(K_{doctor}record \wedge destroy)$, if it is the case that the doctor's knowing the record and destroying the record is a (normal) case for the doctor to know the record $\Box((K_{doctor}record \wedge destroy) \rightarrow K_{doctor}record)$. Yet this condition does not look like a (normal) case in daily life. We will give a further analysis of strong permission in Section~\ref{sec:fcp}.

Since claim-rights are often declared in legal or moral codes, they might be mixed up with the what we called strong permission~\cite{von1968essay}. However, a permission, weak or strong, always concerns our own action, not someone else's. Permission exhibits stronger resemblance to power. In some sense, both refer to some normative possibility of ours. There is a very important difference though in their consequences, as Makinson points out in \cite{Makinson}: if we do something without permission, the consequence will be some punishment. However, if we try to do something without power, we won't be able to do it. As it is argued for in \cite{MarkovichSL}, power concerns actions which are created by constitutive rules (the theory of which is well-known in the presentation of Searle \cite{Searle}). We cannot sell something if it is not ours: selling is a constituted notion (action), the rules describing what happens when someone sells something constitute it. Property on a thing involves a power to sell, if we don't have it, we cannot sell the thing.

\subsection{Privacy, Rights and Permissions to Know: Defeasibility, Co-existence, and Interaction with Knowledge}

The concepts we use in the scope of knowledge and normative notions might influence what modelling we can (or should) use. The right to privacy (see in the Example~\ref{exam:data} of sensitive data) is a complex notion involving different rights and referring an already composite concept (privacy) \cite{privacy}, but its core is a bunch of (claim-)rights towards everyone else to keep away from the private zone, that is, not to get to know about one's private life---which means a long list of prohibitions: practically any action which might end up gathering and disclosing information regarding the private zone.

These rights, like most of the rights (there are a very few exceptions in the modern constitutional democracies) are \emph{defeasible}. This property means that their existence (the truth of the sentences expressing them or the inferences we would draw from them) might depend on circumstances. It might happen that the information that someone's right to know would concern is secret for some reason, and there are stronger arguments keeping it so than supporting the agent's right to know it---in this case, there is no obligation to let them know
. Or the other way around: someone's right to privacy, which would mean that others shouldn't know about their private life and information, can also be defeasible. It can be the case that what they do in their private life might be dangerous for others, and public interest is often a strong argument against the individual one. Sometimes it is far from easy to decide which argument is stronger and then we face a dilemma. Both cases can be represented with using formal argumentation (see Section~\ref{sec:at}).

Consider the interaction between permission to know and knowledge, when we talk about data and inquiry of this data in the context of right to know. As a citizen you are allowed to know the historical records of the city. This sounds a permission, and it seems to be obvious that you therefore can (that is, are permitted to) walk into the archives and have a look. But if you make a data inquiry, then we would agree that the archives should send you back the data you enquired. This means, though, that this allowance you have doesn't only cover a permission but also a claim-right. So we have $P_a[a](inquiry)$ and the form of $R_a[b]K_a(data) \rightarrow O_{b,a}[b]K_a(data)$ holds in this scenario. But such an everyday allowance reveals other challenges using multi-modal logics too, consider now the interaction with knowledge: if we still think knowledge involved here should be classical as ``justified true belief'', namely knowledge implies truth, then we would get to a weird conclusion $O_{b,a}[b]\varphi$ from the premise of right to know $R_aK_a\varphi$: The archives has to see to it that what the the data says is achieved. It is not clear whether knowledge still implies truth, interacting with normative notions. Similar challenges of knowledge occur in the other aspects of social reasoning~\cite{fagin2004reasoning}. One famous case is called the Moore paradox~\cite{van2007dynamic}, which engages the ``pragmatically paradoxical formula'' $K_a(\varphi \wedge \neg K_a\varphi)$. In which sense of knowledge we want to have, when we are modelling our natural language of permission to know? This is an important issue in social reasoning.

\subsection{Conflicts among agents in AI logic}



The beliefs of two agents may be different, and then we can see that the agents conflict. However, there are much more ways in which two agents can be called conflicting. For example, they may try to execute actions with contradictory consequences. Or they may have incompatible rights. The BOID logic~\cite{broersen2002goal} adopts Thomason's idea of prioritization~\cite{thomason2000desires} and gives priority to different types of belief, obligation, desires, and intentions, and then settle conflicts by stating different general policies regarding to the agent type. In the following section we introduce formal argumentation among agents to represent conflicts between logical formulas. 

There are two ways to define conflicts in logic. First we can use a formal semantics, and then there is a conflict if there is no model of the sentences. The semantics of modal logic is based on possible worlds. So for each modal operator there is a set of accessible worlds, and a formula like $O\phi$ is true when $\phi$ is true in all accessible worlds. If the semantics is based on accessible worlds with serial relation, we cannot have a model together for sentences like $O\neg(treat)$ and $O(treat)$. Alternatively, we can define a proof system for modal logic, and then there is a conflict given a set of premises if we can derive a formula and its negation. An additional advantage of a proof system is that the axioms of the proof system reflect the characteristic properties of the logic. 






\section{Argumentation Theory} \label{sec:at}

Argumentation theory analyses the debates in social reasoning, in order to return agreeing from disagree. Debate is quite common in social life. Consider the following case in which three types of conflicts in AI logics are included. We use argumentation theory as a useful tool to represent and (sometimes) resolve conflicts in such realistic situation. In order to illustrate the methodology and tools of formal argumentation, we consider an example regarding the right to know in relation with a globally discussed issue, the question of abortion.

\begin{example}[Abortion]
It is a widely shared thought that the expectant parents have a right to know whether the foetus is expected to be a healthy child after birth, which means that the doctor has to tell them if this is the case. One argument supporting this is that they have a right to decide whether they want to have a baby which is expected to be seriously ill---and in order to decide, they need to know about it. This comes with the abortion being permitted (if it is not permitted, there is nothing to decide about). There is an other, actually contradicting opinion saying that the abortion is very wrong and what supports this is that everyone has a right to life, that is, all the others---also the expectant parents---are forbidden to take this life. This support (the relation between the two latter claim) can be attacked by saying that life as such starts with birth, that is, having an abortion doesn't violates anyone's right to life. In Texas, USA, the senate accepted a bill declaring that parents who had a seriously ill baby born, cannot sue the doctor because of not identifying this illness during the pregnancy. As it is argued in \cite{MarkovichSL} if something (or its compensation) is not subject to enforcement in court, then it is not subject of a duty. This means that in Texas, the expectant parents have no right to that the doctor identify the illness of a foetus. We model identifying as $[d](K_d(ill)\lor K_d(\neg ill))$, that is, the doctor sees to it that he knows whether the foetus is ill. The question is how this action relates to the (most probably) duty-based obligation that if he knows that it is ill then he has to tell to the parents (where telling the parents whether the baby will be ill would be modelled as $[d](K_p(ill)\lor K_p(\neg ill))$). In theory, these seem to be independent: just because the doctor doesn't have to identify, he has to tell if he identifies. But, in practice, as it is now widely discussed (both in Texas and in legal theory), this seems to give the possibility to the doctor to decide: we don't know whether he doesn't identify the illness, or he does but he decides to not tell about it to the parents in the fear of that they might decide to not have the baby. We cannot differentiate between these situations as the access to someone's knowledge seems to be rather exclusive: if the doctor knows about the illness but doesn't let anyone else to know, then we might never know about what he knew or didn't know. This, though, means that not only the identifying is what cannot be sued because of, but also telling the parents if the doctor does know. But if the doctor doesn't have the duty to tell (as it is cannot be enforced) then the question of identifying becomes the question of telling: whether, on the practical level, it also means that the parents don't have a right to know whether their foetus is expected to be healthy or not.
\end{example}

\begin{sidewaysfigure}
	\noindent
	\makebox[\textwidth]{
		\begin{tikzpicture}[scale = .8]


		\node[mystyle] (A) at (-5,-1.5) {\scriptsize * Knowing that the foetus is seriously ill might affect the parents' decision. \\ $A_1 = \Diamond (K_p(ill) \rightarrow [p]\neg(kid))$};
		
		\node[mystyle1] (B) at (1,-1.5) {\scriptsize * Parents have a right to decide whether they want a child (so abortion is permitted to them as an option). \\ $A_2 = P_p [p]\neg(kid)$
		};

		\node[mystyle2] (C) at (-2,2) {\scriptsize Expectant parents have a right to know whether the foetus is expected to be healthy. \\ $A_3 = A_1, A_2 \Mapsto  R_p[d](K_p( ill)\lor K_p(\neg ill))$};

		\node[mystyle] (D) at (-2,6) {\scriptsize Doctor has the duty to tell them. \\ $A_4 = A_3 \mapsto K_d(ill)\rightarrow O_{d \rightarrow p} [d] K_p (ill)$};


		\node[mystyle2] (A') at (7,2) {\scriptsize * Texas Senate Bill 25 (2017): parents giving birth to an unhealthy child cannot sue the doctor because of failing to identify the illness.\\
		$B_1 =  ill\wedge \neg K_d (ill)\wedge\neg R_p (sue_d)$	
		};

		\node[mystyle] (B') at (13,2) {\scriptsize * One has a right to something iff she has the right to enforce it by court in case it's not fulfilled.\\
	$B_2 = R_a [d]\varphi \rightarrow\Box(\neg \varphi\rightarrow R_a (sue_d)) $	
};
		
		\node[mystyle] (C') at (10,6) {\scriptsize The parents don't have a right to that the doctor identify the illness of the foetus\\ $B_3 = B_1, B_2 \mapsto ill \wedge\neg R_p[d](K_d(ill)\lor K_d(\neg ill)$
};

		\node[mystyle] (D') at (7,10) {\scriptsize The doctor doesn't have to tell the parents if the foetus is ill.\\ $B_4= B_3, B_5 \Mapsto K_d(ill) \land \neg O_d[d]K_p(ill)$
};

		\node[mystyle, dashed] (E') at (4,6) {\scriptsize No one can know what the doctor knows if he doesn't let others know about it. \\ $B_5 = \bigwedge_{a\in Agt/{d}} (\neg[d]K_aK_d(ill)\rightarrow K_aK_d(ill))$
		
};


		\node[mystyle] (B'') at (7,-1.5) {\scriptsize Abortion is not okay. (It is forbidden.)\\$C_2 =  C_1 \Mapsto O_p \neg [p]\neg (kid)$
		};

		\node[mystyle] (A'') at (13,-1.5) {\scriptsize * Everyone has a right to life. \\$C_1 = \bigwedge_{a \in Agt} R_a(live)$};

		\node[mystyle] (C'') at (10,-5) {\scriptsize * Life counts from birth (that is, abortion doesn't take one's life) \\$D = \Box([p]\neg(kid) \rightarrow \neg(live))$};


    \draw[dashed] (A) -- (-5, 0) -- (1, 0) -- (B);
    
    \draw[dashed, ->] (-2,0) -- (C);
    
    \draw[<-] (D) -- (C);


    \draw (A') -- (7, 4) -- (13, 4) -- (B');
    \draw[->] (10,4) -- (C'); 

    \draw[dashed] (C') -- (10, 8)--(4, 8)-- (E');
    \draw[dashed, <-] (D') -- (7, 8);


    \draw[dashed, ->] (A'') -- (B'') node[draw=none,fill=none,font=\scriptsize,midway,above] {};


     \draw[postaction={draw,line width=1.5pt,solid}, <->] (D) -- (D');

     \draw[postaction={draw,line width=1.5pt,solid}, <->] (B) -- (B'');

     \draw[postaction={draw,line width=1.5pt,solid}, <-] (10, -1.5) -- (C'');

		\end{tikzpicture}
	}
	\caption{Modelling the abortion example, where we investigate the case when the newborn is seriously ill $(ill)$ and where $[p]\neg(kid)$ means the act of abortion. } \label{fig:abortion}
	
\end{sidewaysfigure}
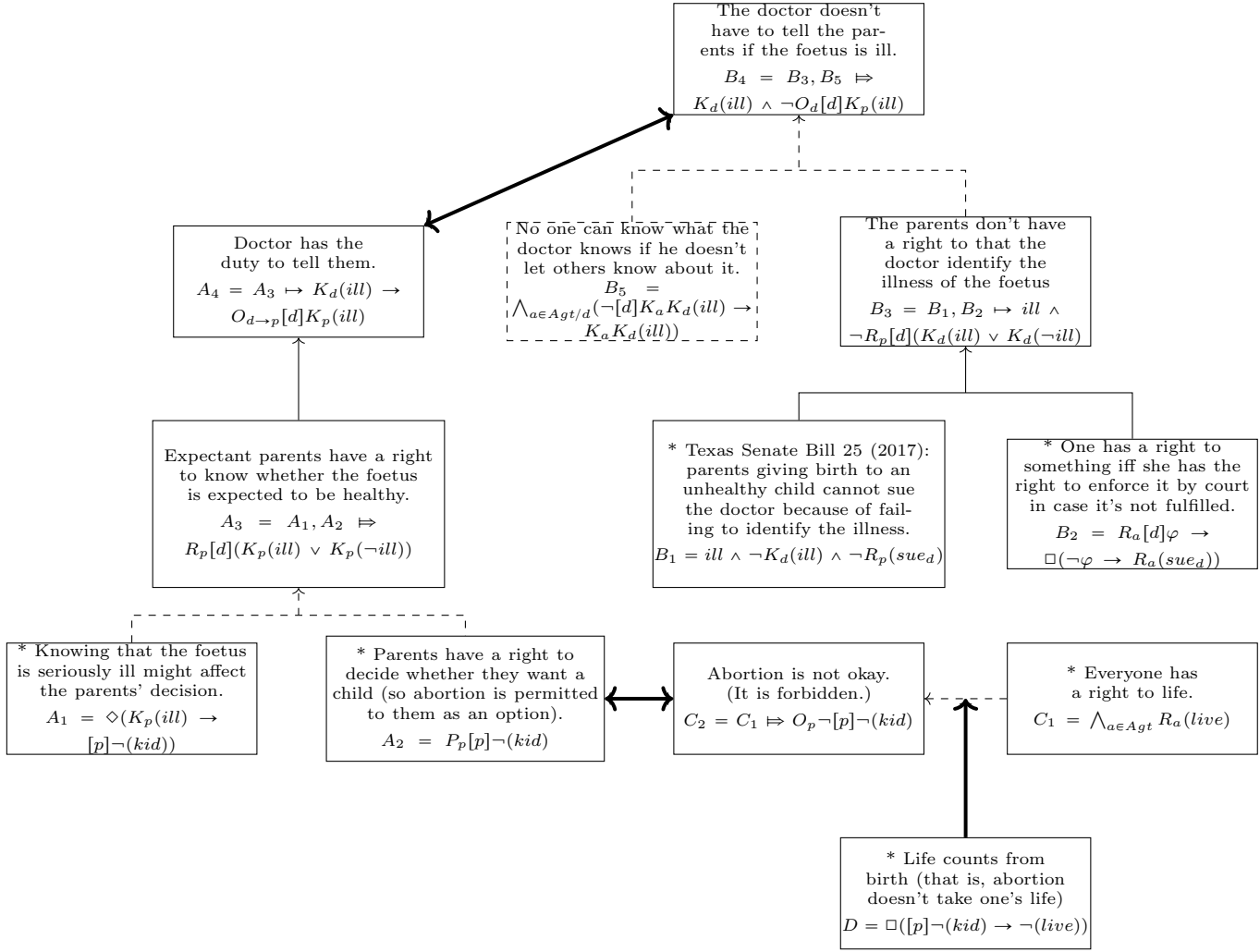

\subsection{Defeasibility}

Reasoning in social reality is often not deductively valid like in modal logics, but rather it is uncertain and wants conflict be resolved. Social reasoning with uncertainty is ubiquitous. The doctor is obliged to treat the patient given the patient is deadly ill $ill \rightarrow O_{doctor}(treat)$. But the doctor is obliged not to do so if she does not know the patient's medical record $\neg O_{doctor}\neg(treat) \rightarrow K_{doctor}(ill)$. So far so good. But when the scenario is that the patient is ill but the privacy of the patient stops the doctor to know the patient's medical record, namely the set of assumptions is $\{ill, \neg K_{doctor}(ill)\}$, then we conclude with $O_{doctor}(treat)$ and $O_{doctor}\neg(treat)$, the doctor's obligation to treat the patient and her obligation not to treat the patient. Which conclusion we should draw? This uncertainty reflects a normative conflict~\cite{ginsberg1994ai,goble2013prima} AI logic wants to resolve~\cite{nute2012defeasible}. 

As argued in~\cite{gabbay1985theoretical,gabbay1994handbook,nute2012defeasible}, a key property of intelligence is to deal with the \emph{uncertainty} of the conclusions we draw in social reasoning, when a new piece of evidence as \emph{exception} is given. The general idea of handling this issue is called ``\emph{defeasibility}'', which will be explained in this section.


\subsection{Formal argumentation to represent and (sometimes) resolve conflicts among agents}



The basic idea of conflict resolution is straightforward: in classical logic, conflicts are inconsistent. If there is an argument that concludes the the door is open, and another argument that the door is closed, then in classical logic everything follows. However, in AI logics, only some of the arguments which are logically valid, will be accepted. Thus, AI logics work as a filter on all possible arguments~\cite{ginsberg1994ai}.

There is one subtlety: formal argumentation does not assume that all conflicts can be resolved. Sometimes there is not enough information to resolve a conflict, or there are simply two incompatible alternatives. In such cases, formal argumentation still plays an important role, because it makes explicit where the disagreement is. This is often expressed by the aim: agree to disagree.

The special property of formal argumentation is that it does not consider the role in which arguments are presented in a dialogue. On the contrary, it assume that all arguments are presented, and it is decided which arguments are accepted, and which ones are rejected. This is inspired by the way the law works: all arguments are presented to the judge, then the judge decides. The order in which the arguments are presented or discovered, is assumed to be irrelevant.

One important property of formal argumentation, as well as other AI logics, is that conclusions can be lost once more information becomes available~\cite{gabbay1994handbook,nute2012defeasible}. This allows for default reasoning. For example, it may be assumed that a bird Tweety flies, until we learn that it is a penguin. In this sense, AI logics are non-monotonic logic~\cite{ginsberg1994ai}.

\subsection{Defeasible knowledge, defeasible rules, and arguments}

Argumentation theory, by comparing the \emph{(de)feasibilities} of arguments systematically, provides a method to clarify the non-monotonic reasons in social reasoning. Generally speaking, argumentation theory defines arguments by either \emph{defeasible} knowledge or \emph{defeasible} inference rules, such that arguments can be attacked on their defeasible knowledge as premises, the conclusions of defeasible rules, and the applications of these rules themselves, in order to explain that the premises of such a rule presumptively rather than deductively support their conclusions~\cite{modgil2018abstract}. In the abortion case it clarifies how the \emph{defeasibilities} are possible in different stages when people debating.

 Two types of knowledge and inference rules are differentiated (See Figure~\ref{fig:abortion}). The ones which are not defeasible then are \emph{strict} (the ones are defeasible are always drawn in dashed). Strict rules (denoted as $\mapsto$, solid links) are intended to capture inferences that are deductively valid, which garantees the truth from premises to the conclusion. While defeasible rules (denoted as $\Mapsto$, dashed links) indicates presumptive inferences, where the premises likely supports that conclusion, but which can be refuted if a evidence to the contrary is given. Arguments are built from all knowledge by applying rules recursively. An argument is \emph{defeasible} if the rule it applied is defeasible; otherwise \emph{strict}. An argument is \emph{firm} when all its knowledge in the premises are strict; otherwise \emph{plausible}. In Figure~\ref{fig:abortion}, the boxes with $*$-marks are knowledge, the ones without are the conclusions of arguments given by some inference rules. The argument for ``Abortion is not okay'' is defeasible because the rule it applied, while the argument ``Life counts from birth'', as a strict knowledge, is strict. In Figure~\ref{fig:abortion} each block represents a claim in the debate. They are the conclusions of arguments. Argument $C_2$ is defeasible while argument $D$ is strict. All arguments are firm here except $B_4$ and $B_5$.     

\subsection{Preferences among Arguments}


Here we first introduce the rule-based preferece~\cite{dong2019aspic} to give arguments priority. 
The rule-based preference always preferes strict arguments than the defeasible ones. Thus this preference (strictly) prefers $B_3$ than $A_4$ in Figure~\ref{fig:doctor}.
There also other ways of comparing arguments. One is called universal preference, another is called premise-based preference~\cite{modgil2018abstract,dong2019aspic}. The universal preference prefers every argument the same. Thus the universal preferenc is an equivalent relation over arguments. We can also catergorize arguments by the property of their premises. The premise-based preferenc gives priority to the firm arguments than the plausible ones. 


\subsection{Resolving conflicts using defeats among Arguments}

Taking the arguments with contraries together it costs inconsistency, and then we can compare one with another by the notion of defeat. We follow after~\cite{modgil2018abstract} and define three kinds of defeats: rebuttal, undermining, and undercutting. In the debate of abortion this distinction is well illustrated. The first and the most well-known defeat relation is rebuttal, which considers the defeasibility based on the conclusions of inference rules. In this debating, the conclusion ``Doctor has the duty to tell them'' of strict argument $A_4$ is contrary to the one of defeasible argument $B_4$. As $A_4$ is more preferable than $B_4$ in the rule-based preference, $A_4$ rebuts $B_4$. The second notion, undermining, concerns the defeasibility on the premises of the arguments. The premise ``Parents have a right to decide whether they want a child'' of firm argument $A_2$ is contrary to the conclusion ``Abortion is not okay'' of firm argument $C_2$, and also $A_2$ is not preferable than $C_2$ in the premise-based preference, then we say $A_2$ is undermined by $C_2$. The last one the undercutting specifies how one argument defeat another based on the application of inference rules. Based on the rule-based preference, the conclusion ``Life counts from birth'' of strict argument $D$ contradicts the inference rule which is applied by defeasible argument $C_2$ for ``Abortion is not okay'', and thus $D$ undercuts $C_2$.

\subsection{Stable Extensions} \label{sec:extension}

How to select a set of justified arguments without no conflict from the one with? In 1995 Dung~\cite{dung1995acceptability} proposed a notion called ``stable extension'' to deal with conflict. In Dung's idea, a set of argument is a stable extension when all argument defeated by this extension do not belong to it. In other words, at first no argument contained defeat the ``insiders'', and, secondly, all ``outsiders'' are defeated by some argument in the extension. Each stable extension contains the claims which are accepted together. Thus a stable extension provides a sense of consistency to explain when a set of arguments has no conflict. In Example~\ref{exam:abortion}, the arguments for the claim ``Doctor has the duty to tell the parents if the foetus is excepted to be ill'' forms a stable extension, saying arguments $A_1, A_2, A_3, A_4, B_1, D$ together they accept each other.

\section{Freedom To Act} \label{sec:fcp}

As explained in Section~\ref{sec:permission}, if I tell you that you are permitted to have green or black tea, I suggest to you that you are free to choose whether you would like to have green or black tea. Consequently, you are permitted to have black tea and you are permitted to have green tea. This is known as free-choice permission~\cite{vonWright1963,hansson13:_variet_permis}. In this section we consider free-choice permission in the context of AI logic of knowledge and action. For example, assume you permitted to know the address or to know the phone number, does it imply that you are permitted to know the address and you are permitted to know the phone number? And if you are permitted to walk or take the bus, are you permitted to the action of walking and are you permitted to the action of taking the bus?

In particular, in this section we show how argumentation-based AI logic can deal with some of the challenges of free choice permission using formal argumentation. 
It is well known that free-choice permission has to be treated with care.
For example, the ``free choice paradox''~\cite{hansson13:_variet_permis} indicates one core of moral conflict, which is led by an inappropriate way to put obligation, permission, and possible acts together. 
It is so explained by the two principles FCP and OWP in deontic logic~\cite{van1979minimal}. 

As illustrated in Section~\ref{sec:permission}, FCP captures the intuition of ``freedom'' or ``act liberty'', characterizes free choice permission~\cite{hansson13:_variet_permis} or strong permission~\cite{ano2015symbolic}. 
One instance of FCP occurs as a permission to know in the case of product purchase. To provide the online purchase service, the company is permitted to know the information the customer gives or to know the information about the customer use. This concludes that the company has the permissions to know these two types of information. This ``freedom to know'' for the company is seen as a necessary part of providing service, written in its privacy notice. 

OWP expresses that all permissions are put into the shoes of obligations. 
In the case of online purchase, all permissions of knowing different types of personal information for the company are limited by the duties and the obligations it should fulfill regarding to the privacy notice, the privacy shield framework it participates, and the data protection laws. Both principles FCP and OWP are used commonly in different aspects of social reasoning, including reasoning in natural language~\cite{asher2005free}, game theory~\cite{ano2015symbolic,dong2015three}, and legal reasoning~\cite{dong2017dynamic}. 

\begin{figure}
		\noindent
	\makebox[\textwidth]{
\begin{tikzpicture}[scale = .8]

\node[mystyle] (A) at (5,-1.5) {* It is permitted to know the buyer's personal information.\\ $B = P_cK_c(customer)$};
\node[mystyle] (B) at (0,-1.5) {* It is forbidden to misuse the personal information. \\$A = O_c\neg (misuse)$};
\node[mystyle2] (C) at (2.5,2) {Knowing personal information does (normally) not bring about misusin.\\$C' = A, B \Mapsto \Box(K_c(customer) \rightarrow \neg (misuse))$ };
\node[mystyle3] (D) at (-6,2) {* It is able to misuse personal information for other purposes. \\$C = \Diamond(K_c(customer) \wedge misuse)$};
\node[mystyle3] (E) at (5,8) {It is permitted to know personal information to take and handle orders.\\$B'' = B \Mapsto P_c(K_c(customer) \wedge handle)$};
\node[mystyle3] (F) at (-7,5) {It is not permitted to know personal information for other purposes.\\$A'' = A, C \Mapsto \neg P_c(K_c(customer) \wedge misuse)$};
\node[mystyle3] (H) at (1,5) {It is permitted to knowing personal information for other purposes.\\$B' = B \Mapsto  P_c(K_c(customer) \wedge misuse)$};

\draw[dashed] (B) -- (0, 0) -- (5, 0) -- (A);
\draw[dashed] (2.5, 0) -- (C) node[draw=none,fill=none,font=\scriptsize,midway,right] {OWP};

\draw[dashed] (-1, -.5) -- (-1, 3.4) -- (-6, 3.4) -- (D);
\draw[dashed] (-5,3.4)-- (-5,4) node[draw=none,fill=none,font=\scriptsize,midway,right] {OWP};

\draw[dashed] (6,-.5) -- (6,6.8) node[draw=none,fill=none,font=\scriptsize,midway,right] {FCP};



\draw[dashed] (H) --node[draw=none,fill=none,font=\scriptsize,midway,above] {FCP} (5.5,5) -- (5.5, -.5);
\draw[postaction={draw,line width=1.5pt,solid}, ->] (F) -- (H);
\draw[postaction={draw,line width=1.5pt,solid}, <-] (C) -- (D);

\end{tikzpicture}
}
\caption{The case of knowing the customer's personal information. } \label{fig:knife}
\end{figure}
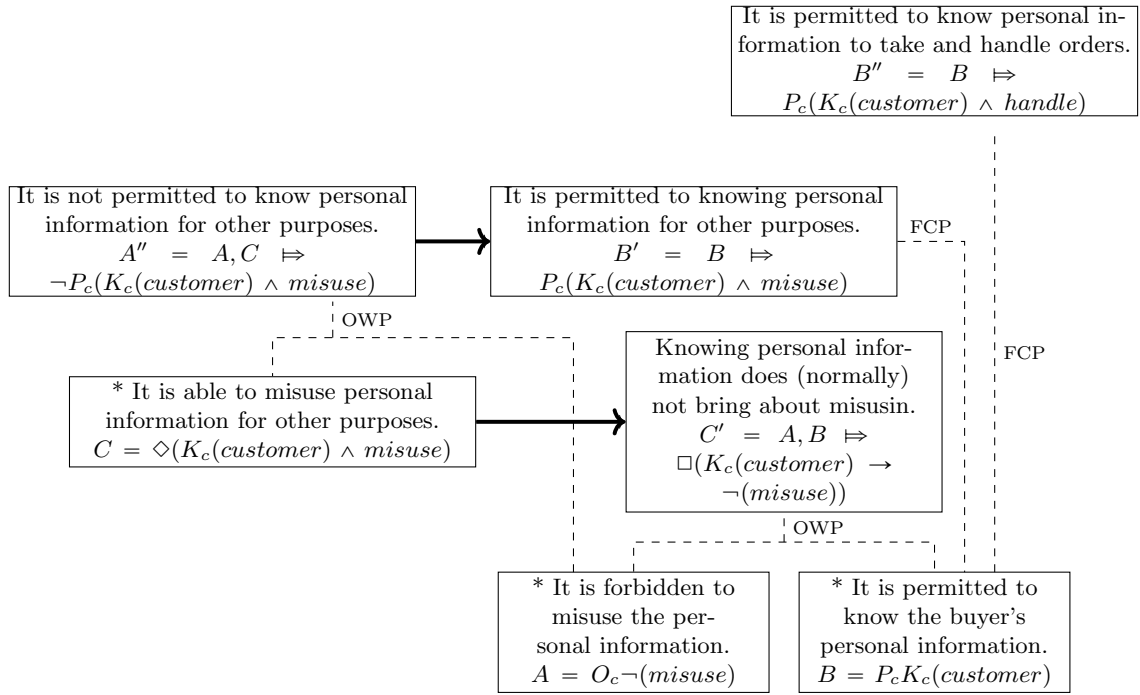

The two principles are consistent with each other, but might not when together with some \emph{exceptional assumptions}. Consider the following example in Figure~\ref{fig:knife}:  
\begin{enumerate}
	\item It is permitted to know the personal information from the customers.  ($B$) 
	\item It is forbidden to misuse the information for other purposes. ($A$)
\end{enumerate} 
Adopting the ``obligation as the weakest permission'' principle, we then receive a conclusion that
\begin{enumerate}
	\setcounter{enumi}{2}
	\item Knowing personal information (normally) does not lead to misuse. ($C'$)
\end{enumerate}
However, we also have the \emph{exception} ``Knowing the personal information is able to misuse for other purposes'' as a premis ($C$), which is a contradiction of the previous statement. We \emph{prefer} this new information of possible action than the statement $C'$ concluded previously, as this new piece is an evidence as a social fact while the old one is defeasible. We then no longer expect to derive that knowing personal information are (normally) performing the purposes of service ($C'$). Further, we neither infer that misusing the information is permitted ($B'$), because which is a conclusion from applying the defeasible FCP on free chioce permission of knowing personal information. In this social scenario, putting permission into the shoes of obligation is more plausible and preferable than having the defeasible rule to always behave freely. Therefore OWP overrides FCP. However, due to our social mind we still would like to expect that, for instance, the company is permitted to know personal information to take and handle orders ($B''$), because there is \emph{no more preferable} argument for the contrary.

\section{Research program for social AI logic} \label{sec:futurework}

The breath-taking development of AI is fuelled by the increase in computing power in high performance computing and other computing technologies. Most of the techniques now deployed in machine learning have been around for a several decades, but only with current computing technologies they could be deployed widely. It is broadly assumed that similar revolutions can also be expected in other areas of AI, such as representation and reasoning. In particular, whereas data science and machine learning have given rise to what has been called animal AI
, it is expected that developments in information and knowledge technology will lead to human AI, though it remains an open debate what this human AI will look like. 

For the development of human AI logic, we need to understand the challenges of knowledge representation and reasoning. Several lessons can be learned from the study of expert systems in the eighties, which came with many promises, but which were missing the computing power we have today.
In this section we use the discussion in the four research questions in this paper to develop a research program for the future development of argumentation-based social AI.


The starting point of our research program is the handbook of logic for artificial intelligence and logic programming~\cite{handbook_ailp}, written thirty years ago. In those days, logic for AI contained five topics: logical foundations, deduction methodologies,  non-monotonic reasoning and uncertain reasoning, epistemic and temporal reasoning, and logic programming.  During the past three decades, we have witnessed the following developments in these areas:
\begin{enumerate}
\item
Fragments of first-order logic have been developed in much detail, in particular description logics to represent and reason about ontologies.
\item
Existing deduction methodologies have been adapted to new computing technologies, and interactive theorem provers like Isabelle have become much more user friendly \cite{DBLP:journals/corr/abs-1903-10187}.
\item
The dominant approach in dealing with non-monotonic reasoning and uncertainty is formal argumentation, with its COMMA conference series. The first volume of the handbook series of formal argumentation appeared last year, and the second volume is in preparation.
\item
Epistemic and temporal reasoning has been extended to agent reasoning, involving also actions, preferences, intentions and so on. It is discussed at the AAMAS conference series.
\item Logic programming has evolved into answer set programming~\cite{handbook_ailp}.
\end{enumerate}
Moreover, as explained in this paper, the current challenge is the application of AI logic to social reasoning.


\subsection{Argumentation-based logic}

The first step of our methodology is to bring formal argumentation and formal logic together by developing argumentation based logic. We do not distinguish here between theoretical and practical reasoning.

In the current paper we assumed that a formal argumentation theory like the \emph{rule-based} argumentation theory ASPIC$^+$. It can adopt two base logics, the lower-bounded logic and the upper-bounded logic, ``one for the strict and one for the defeasible rules''~\cite{modgil2018abstract}. They call this the crude approach to define arguments. Basically, chosen two modal logics, strict rules can be defined by the derivations in the lower-bounded logic, while defeasible rules are by those derived in the upper-bounded logic but not the lower-bounded one~\cite{dong2019aspic}. This methodology, as argued by Modgil and Prakken~\cite{modgil2018abstract}, is based on certain cognitional or rational criteria. This method can be generally used to define argumentation theory and, further, instantiates different non-monotonic AI logics based on this systematical theory for social reasoning. 

A further research along this line is to adopt the distinction for knowledge and then to differentiate arguments accordingly~\cite{modgil2018abstract}. We have distinguished knowledge into \emph{fallible} and \emph{infallible}. In this case, the \emph{fallibility} of an argument need not only be located in its premises,
but can also be located in the inference steps from premises to conclusion~\cite{modgil2018abstract}. This distinction provides a different preference a premise-based preference to compare arguments, and thus defines different kinds of defeat to resolve moral conflicts. Such distinction will be reflected in the constructions of AI logics by differentiating derivations with different kinds of premises. 



We have suggested elsewhere~\cite{dong2019aspic} how to provide a general method to develop AI logics based on stable extensions regarding to the two monotonic logics and the rule-based order. It is a step-by-step construction similar to the proof of Lindenbaum Lemma  in modal logic~\cite{blackburn2002modal}. The basic idea is to construct a set of consistent arguments based on the lower-bounded logic first and then the upper-bounded logic second, in a way of balancing the arguments given by strict and defeasible rules to preserve maximal consistency. 

Our AI logics are defined based on the stable extensions constructed by this two-step construction in a step-by-step mechanism~\cite{dong2019aspic}. All logical inferences are either from the lower-bounded logic or from the upper-bounded logic. 
We distinguish three steps. First the development of argumentation-based logic, second the development of argumentation-based AI logic, and finally the development of argumentation-based social AI logic.

\subsection{Argumentation-based AI logic and reasoning}

In the second step of our research program, we focus on the particularities of practical reasoning in AI logic. Traditionally theoretical reasoning was assumed to be easier than practical reasoning due to correspondence between the theory and observable reality. However, with the rise of modern physics such as quantum mechanics, doing experimentally testing theories has become harder as well. Nevertheless, in general it is more difficult to formalize practical reasoning than theoretical reasoning due to the inherent ambiguity in practical reasoning and the lack of expirmental data. The examples in this paper suggest the following.
\begin{description}
\item[COMPLEXITY.] Formalization is difficult, as shown by the examples in this paper, where several formalizations of each sentence were given, each with their advantages and disadvantages.~\footnote{In Example~\ref{exam:data}, there are three ways to formalize the firm knowledge ``The doctor has an obligation to treat an ill person.'' $O_d \langle d\rangle(treat)$ is the one we presented in Figure~\ref{fig:doctor}, while the other two can be: $O_d [d](treat)$ and $ill \rightarrow O_d [d](treat)$. } 
\item[MULTIPLICITY.] The ``best'' formalization often does not exist, there is often a trade-off between representations or models. 
\item[ARGUMENT ELICITATION.] 
People may misrepresent their argument to various degrees, and we need a methodology in which people can adjust their arguments. 
\end{description}

The current data driven approaches (natural language processing, argument mining, machine learning) will help us in the above task of argument elicitation, and more generally in the task to design better social AI logics.

\subsection{Argumentation-based social AI logic}

Not only the representation of examples must be flexible, also more generally the right language for social AI logic, the right axioms and the right semantics, need to be developed experimentally. This explains the ``Towards'' in the title of the paper.

Consider strong permission which plays a role of representing trust in multi-agent interaction. Trust reflects multi-relationship of dependency in social interaction. It provides freedom for the trustworthy individual to react. On the other hand, the trustworthy institutions are usually required to bring positive consequences in most of their activities. Thus, to gain the others' trust in order to act at certain liberty, the trusted entities need to have better strategic plan for their behaviors. The formal models of these social phenomenon are needed to figure out the further connection between trust, permission, and strategic actions.

\section{Related Work} \label{sec:related}

There are various concepts from AI logics as well as from social reasoning which are studied in the literature, but which we do not consider here.
The general area of AI logic and social reasoning is huge~\cite{boella2006introduction}. Some often discussed concepts are listed in Table~1 below. In this paper, we restrict our scope to the concepts necessary to analyze arguments around the right-to-know and privacy. We therefore do not explicitly discuss trust, reputation, responsibility, liability and so on. However, our AI logics can be extended to deal with these concepts as well.
\begin{table}[ht]
	\centering
\begin{tabular}{ r|c|c| }
	\multicolumn{1}{r}{}
	&  \multicolumn{1}{c}{AI/Individual}
	& \multicolumn{1}{c}{Social/Collective} \\
	\cline{2-3}
Obligation & Ought & Responsibility, Liability \\
	\cline{2-3}
Permission & Freedom & Right (to use, to know, to privacy), Power \\
	\cline{2-3}
Knowledge & Uncertainty, Planning & Trust, Fairness, Bias \\
	\cline{2-3}
Action & Causality & Social Dependency, Institution \\
	\cline{2-3}
\end{tabular}
\vspace{10pt}
\caption{\label{tab:concept}Concepts regarding individual and collective reasoning.}
\end{table}

In social reasoning, the impact of the behavior of other agents on an agent is considered, and social dependencies among agents are made explicit~\cite{horty2001agency,boella2006introduction}.  Moreover, social dependences are related to trust among agents~\cite{castelfranchi1998normative,saam1999simulating}. On the one hand, trust is a challenge since our trust can be violated, leading to negative consequences for our individual planning~\cite{falcone2001social}. On the other hand, we can achieve much more by depending on other agents, then we could achieve by ourselves~\cite{castelfranchi1998modelling}. From this perspective, social dependence and social trust are the glue of society. In epistemic logic, they are represented as beliefs and trusts in social networks~\cite{seligman2011logic,liu2014logical,liau2003belief}. In normative reasoning, social networks are represented by all the rights among agents. 

Some social norms are institutionalized in the law~\cite{kanger1966rights,kanger1970new,lindahl1977position,jones1996formal,grossi2008many,sergot13_normative}. In the same way, the law may be seen as a way to constrain agents, or as a mechanism to empower them~\cite{jones1996formal,lindahl1977position,sergot13_normative}. On one hand privacy norms may constrain what agents can know about other agents, but on the other hand it helps the agents to protect their own data and to maintain their privacy~\cite{aucher2011dynamic}. 


It is common practice that also other cognitive attitudes of agents are modelled in the same way, such as their desires, goals and intentions~\cite{lang2003hidden,horty2001agency,benthem2014priority}. Moreover, special connectives have been developed for action and change~\cite{van1995language}, for example to express causality. In this paper we do not consider this, but instead we present three topics in related as follows. Comparing to what they have been done, our AI logic for social reasoning is a DEAL: deontic epstemic action logic.


{\bf Logics of Obligation and Agency} One widely applied logic of obligation and agency is the STIT-logic~\cite{horty2001agency}. STIT-logic offers a BDI framework to illustrate agentive obligations, based on Bratman's philosohpy about practical reasoning~\cite{meyer2015bdi}. The semantics of STIT-logic is a temporal branching-time model which can fine-grain agents' choices: each choice of an agent is depended on the moment and history the agent takes. Agents' actions are implicitly expressed by the consequences the actions of the agents bringing about. This is the basic idea of STIT-modality: Agent $a$ sees to it that $\varphi$ ($[a: \sf STIT]\varphi$) iff agent's choice ensures $\varphi$. So, rather than saying ``ought to do'' STIT-logic says ``ought to be'' regarding to agencies: What is obliged for an agent is that all ideal choices of this agent are seeing to it that. Indeed STIT-logic not only well captures individual actions but also collective ones~\cite{lorini2013temporal}, and thus it has be widely applied into the analysis of social influence~\cite{lorini2016stit}, strategic games~\cite{tamminga2013deontic}, and legal reasoning~\cite{sergot13_normative,MarkovichSL}. Recently STIT-logics of obligation and knowledge have been developed~\cite{pacuit2006logic,horty2019epistemic}. For more discussions of STIT please refer to~\cite{pigozzi2017multiagent}.

{\bf Logics of Conditional Norms} Conditional norms usually express statements regarding to deontic detachment like ``$\varphi$ is obligatory if $\psi$''. That is to say, given a conditional norm and its antecedent, what deontic consequent of the conditional statement is inferred. Deontic logicians explored different formal accounts to capture this feature generally, e.g. input/output Logic~\cite{parent2013input} as well as preference-based deontic logics~\cite{hansson2013alternative} and their dynamics~\cite{benthem2014priority} are two main accounts for the process of detachment. In application, the logics of conditional norms can be used to discuss constitutive norms in legal theory~\cite{grossi2008many}. 


{\bf Logics of Rights and Permissions} To fully consider the interaction between agents by modelling their normative behaviors, permission is one very important but less explored concept of norms in social reasoning~\cite{hansson13:_variet_permis}. 
Permission is often seen in social life. ``This journal is open accessed. So everyone is allowed to read and download the articles freely''
, ``After validate the full monthly pass, you are permitted to take whatever public transport in the Luxembourg traffic networks in a month''
, ``Signing the real estate purchase contract, you are allowed to freely decorate your own house''
. It involves the topics of freedom, privacy, empower etc. which cover nearly all areas in our social life. To understand what a given permission means, we need a proper way to reason about this permission and its relations toward to this agent's obligation, knowledge, and what behaviors he or she is able to execute.

A number of researchers, in the area of philosophy, linguistics, computer science, and legal theory, have pointed out the many faces of permission. Permission, for instance, can be seen as weak permission as the dual of obligation~\cite{vonWright1951}, strong permission as freedom to act~\cite{vonWright1963,dignum1996free,asher2005free}, dynamic permission~\cite{IOLpermission}, and defeasible permission as exception of existing norms~\cite{hart1948ascription,raz1994ethics}, covering many important rational principles used in multi-agent interaction. 

Here we mainly compare the notions of weak and strong permission, following the idea given by von Wright~\cite{vonWright1963}: ``An act will be said to be permitted in the weak sense if it is not forbidden; and it will be said to be permitted in the strong sense if it is not forbidden but subject to norm. Acts which are strongly permitted are thus also weakly permitted, but not necessarily vice versa.'' One difference between these two permissions is indicated by their interactions with knowledge. Given that you are having a full monthly pass, you are permitted to take any public transport
. However, if it were a weak permission, without receiving that it is not forbidden to not active your ticket before taking the transport, you have no idea whether it is permitted to not active the ticket before entering a train
. The given weak permission cannot guide your action. Yet, by having this pass as a strong permission, when you get the knowledge that entering any transport without the regular activation is an instance of having this pass, you understand that you are permitted to do so
. Further more, in the multi-agent context, your fellow's trust provides you a freedom to stay in her property and use the facilities freely, without any non-prohibition is announced.

\section{Summary}



We present a research program for the logic of social AI, which extends approaches of AI using only data science and machine learning. AI logic formalizes the reasoning of intelligent agents. The logic of social AI extends AI logic with mechanisms for social reasoning. In this paper, we show how an argumentation-based AI logic of action, knowledge, right, obligation and permission can be used to formalize important aspects of social reasoning. Besides reasoning about the knowledge and actions of individual agents, social AI logic reasons also about social dependencies among agents. 

First, we show how rights represent relations between the obligations and permissions of agents. 
For instance, if taking freedom as one form of permission, then one's freedom of knowing something is the others' no claim-right about this knowing toward  this person. While one has a claim-right of knowing the data toward the others, it indicates that the others have duties to let this person know the data. In social practice, this type of interaction between obligation and permission involves agent-dependency.

Second, we show how some conflicts among agents can be identified and (sometimes) resolved by comparing logical arguments about permissions and prohibitions. This is done by applying either the formal logics or the formal tool argumentation theory. The concept of inconsistency in logic is a way to identify conflicts in a debate. But to resolve conflicts we apply argumentation theory. The occurrence of conflict is pointed out by the arguments been defeated. A way of conflict resolution in argumentation theory is to take the stable extensions, in which arguments are all accepted together.

Third, we show how to argue about the right to know, one central issue in the recent discussion of privacy and ethics. 

Fourth, we discuss the issue of free-choice permission in this setting. Rather than simply to say whether free choice permission is a rational concept or not, we presented a pratical case of right to know personal information, and pointed out the issue free choice permission involved can be reinterpreted together with the relation between obligation, permission, and actions. The freedom to use customers' personal information is binded by the legal requirements. Without considering this interaction, it is hard to say the freedom we have is rationally given.

\section*{Acknowledgement}
Huimin Dong is supported by the China Postdoctoral Science Foundation funded project [No. 2018M632494], the National Social Science Fund of China [No. 17ZDA026], and the National Science Centre of Poland [No. UMO-2017/26/M/HS1/01092]. Leendert van der Torre and R\'{e}ka Markovich have received funding from the European Union's Horizon 2020 research and innovation programme under the Marie Sklodowska-Curie grant agreement No 690974 for the project ``MIREL: MIning and REasoning with Legal texts''.


%
%
%
 \bibliographystyle{unsrt}
\bibliography{zju.bib}



\end{document}